\documentclass[sigconf]{acmart}
\acmYear{2025}
\acmConference[arXiv]{February}{February}{2025}
\acmISBN{$^{*}$Corresponding author: arissa@google.com}
\setcopyright{none}
\copyrightyear{2025}
\usepackage{algorithmic}
\usepackage{textcomp}
\usepackage{fancyhdr}
\usepackage{hyperref}
\usepackage{multirow}
\usepackage{graphicx}  %
\usepackage{multirow}  %
\usepackage{array}     %
\usepackage{booktabs}  %
\usepackage{amsmath}
\usepackage{circledsteps}
\usepackage{xspace}

\usepackage[font=footnotesize]{caption}

\hypersetup{
    colorlinks=true,
    linkcolor=blue,
    filecolor=magenta,      
    urlcolor=cyan,
    citecolor=blue,
    anchorcolor=blue,
    pdftitle={Overleaf Example},
    pdfpagemode=FullScreen,
}
\usepackage{microtype}
\usepackage{subfigure}

\AtBeginDocument{%
  }

\begin{document}

\title{Machine Learning Fleet Efficiency: Analyzing and Optimizing Large-Scale Google TPU Systems with ML Productivity Goodput\\[.25em] \subtitle{\emph{A Preprint}}}

\author{\Large
    Arissa Wongpanich$^{\dagger}$
    Tayo Oguntebi$^{\dagger}$
    Jose Baiocchi Paredes$^{\dagger}$\\
    Yu Emma Wang$^{\dagger}$
    Phitchaya Mangpo Phothilimthana$^{\dagger}$
    Ritwika Mitra$^{\dagger}$\\
    Zongwei Zhou$^{\dagger}$
    Naveen Kumar$^{\dagger}$
    Vijay Janapa Reddi$^{\dagger\ddagger}$ \\[1em]
    $^{\dagger}$Google \hspace{1cm} $^{\ddagger}$Harvard University\\[1em]
}

  \renewcommand{\shortauthors}{}

\begin{abstract}
Machine learning (ML) infrastructures operating at warehouse scale present unique performance characterization challenges beyond traditional high-performance computing metrics. This paper introduces a systematic framework for analyzing ML fleet efficiency, demonstrated on Google's production TPU infrastructure comprising thousands of accelerators running diverse workloads. Our fleet-wide analysis reveals performance dependencies spanning the entire ML system stack, from hardware to model architecture, data pipelines, frameworks, compilers, and schedulers. We identify critical gaps in conventional utilization-based performance metrics and propose "ML Productivity Goodput" (MPG) to capture fleet-wide efficiency across heterogeneous ML environments. Using MPG, we provide different methods for identifying and addressing bottlenecks throughout the ML system stack. When applied to Google's TPU workloads that are used in production, our results show that the methodology yields substantial performance improvements, and establishes a basis for efficient management of large-scale ML computing infrastructure.

\end{abstract}

\newcommand{\mpg}{ML Productivity Goodput\xspace}
\newcommand{\sg}{Scheduling Goodput\xspace}
\newcommand{\pg}{Program Goodput\xspace}
\newcommand{\rg}{Runtime Goodput\xspace}
\newcommand{\google}{Google\xspace}
\maketitle
\pagestyle{plain}

\section{Introduction}\label{sec:intro}

\renewcommand*{\arraystretch}{1.25}
\begin{table*}[t!]
\centering
\caption{Comparison of Machine Learning (ML) Fleet, Warehouse Scale Computer (WSC), and High-performance Computing (HPC).}
\label{tab:comparison}
\resizebox{\textwidth}{!}{%
\scriptsize
\begin{tabular}{@{}lp{4.5cm}p{4.5cm}p{4.5cm}@{}}
\toprule
\textbf{Category} & \textbf{Warehouse Scale Computer} & \textbf{High-Performance Computing} &  \textbf{Machine Learning Fleet} \\ \midrule

\textbf{Workload types} & Diverse web services  (search, email, social networking, media streaming) & Scientific simulations, graph computations, solvers & Training of ML models, real-time serving, bulk inference \\ \midrule

\textbf{Fleet composition} & More stable, as most user demand has reached a steady state or known patterns & A large portion of demand is predetermined, as it is driven by scientific missions & Rapidly changing due to newly emerging ML models and increasing user demand
 \\ \midrule
\textbf{Hardware heterogeneity} & General-purpose CPUs
& CPUs, GPUs, other ASICs & CPUs, GPUs, TPUs, FPGAs, other ASICs.\\ \midrule

\textbf{Hardware/Software co-design} & Hardware is workload-agnostic & Hardware is chosen for specialized applications & ASICs often co-designed with workloads
in mind\\ \bottomrule
\end{tabular}%
}
\end{table*}
\renewcommand*{\arraystretch}{1}

As machine learning (ML) models become larger and more complex, production fleets must deploy highly parallel training \cite{vaswani2017attention} and low-latency inference \cite{pope2023efficiently} applications at unprecedented scale. Much like how the boom of internet-scale services \cite{hamilton2007iss} prompted the development of the warehouse-scale computer (WSC), \citep{barroso2009wsc,kanev2015profiling}, the current explosion of foundation ML models \citep{geminiteam2024geminifamilyhighlycapable,brown2020languagemodelsfewshotlearners,anthropic2024claude} is ushering in the era of the ML fleet. ML fleets, sometimes referred to as ``AI hypercomputers'' \cite{aihypercomputer}, represent a new paradigm in computer architecture, where domain-specific accelerators (DSAs) work in tandem with a modular ML system stack to achieve significant performance and energy efficiency improvements. They are characterized by their massive scale and ML-centric workloads, often requiring more compute power, more storage, and more complex networking than traditional WSCs.  Despite the rapid development of these ML fleets, the challenges associated with building and operating them at scale remain significant and poorly understood in the literature.

\begin{figure}[t]
    \centering
    \includegraphics[width=\columnwidth]{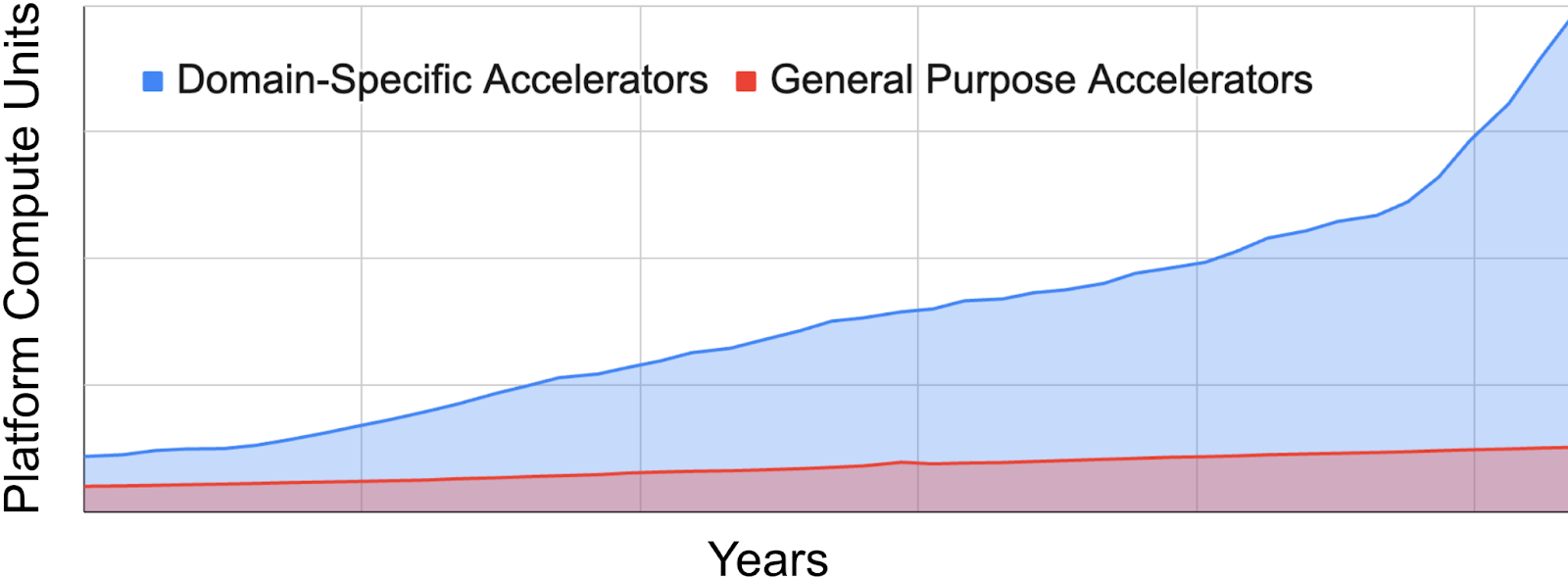}
    \caption{Five-year historical ML fleet breakdown by accelerator type. The rapid proliferation of domain-specific accelerators in response to ML-based workloads has presented novel challenges in optimizing ML fleets. Managing these domain-specific accelerators means effectively handling hardware and workload heterogeneity, as well as hardware-software co-design at scale.}
    \label{fig:five-years}
\end{figure}

This paper presents a playbook for instrumenting and optimizing large-scale ML fleets, using Google's production Tensor Processing Unit (TPU) infrastructure for internal workloads as our case study. In contrast to prior work, where previous Google TPU papers have predominantly focused on hardware architecture design and chip-level performance evaluations \citep{jouppi2023tpuv4opticallyreconfigurable, jouppi2021ten, jouppi2017indatacenter}, our work provides the first in-depth perspective on the software stack that enables these domain-specific accelerators to operate efficiently at scale. 

This shift from hardware-centric discussions to the practical aspects of operationalization is important in understanding how to leverage and manage ML infrastructure in real-world, production environments. Although our study centers on Google's ML fleet, which is primarily composed of TPUs, the insights, methodologies, and best practices we present are broadly applicable to ML fleets consisting of other DSAs such as GPUs.

The contributions of this paper are listed as follows. \Circled{1} First, we provide a methodology to dissect the ML fleet by segmenting it along the layers of a system stack, from the low-level hardware layer to the user-facing application layer. This analysis reveals three major challenges for ML fleet optimization: hardware heterogeneity, workload heterogeneity, and hardware/software co-design. Traditional optimization strategies, which often rely on stable hardware and workload characteristics, are inadequate for addressing these dynamic interactions. Architecture-centric metrics, such as TOPs/Watt or peak FLOPS, while valuable for specific hardware evaluations, fall short in capturing the overall complexity and efficiency of an ML fleet. 

\begin{figure}[t]
    \centering
    \vspace{.5em}
    \includegraphics[trim=0 10 0 0, clip, width=.9\columnwidth]{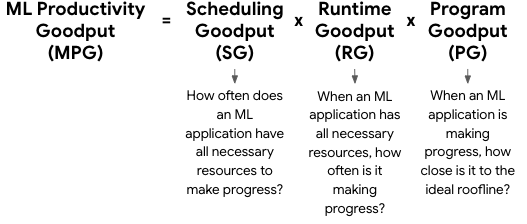}
    \caption{ML Productivity Goodput (MPG) and its components.%
    }
    \vspace*{-1em}
    \label{fig:ironlaw}
\end{figure}

\Circled{2} Our second contribution addresses these issues. To view the fleet holistically and capture its performance across multiple layers of the ML system stack, we introduce the ``ML Productivity Goodput'' (MPG) metric. Unlike traditional metrics, MPG encompasses scheduling efficiency, runtime efficiency, and compiler / program efficiency. MPG is analogous to the ``iron law'' of computer performance~\cite{eeckhout2010computer}, but adapted for ML fleets as shown in \autoref{fig:ironlaw}. The MPG metric enables us to identify areas for improvement across the entire machine learning fleet by analyzing performance across different segments, such as accelerator type, model architecture, and workload phase (e.g., training vs. serving). By breaking down the metric into subcomponents---Scheduling Goodput (SG), Runtime Goodput (RG), and Program Goodput (PG)---and examining it along the axes of fleet characteristics, we can precisely pinpoint optimization opportunities.

\Circled{3} Finally, we provide a procedure for leveraging the MPG metric and show strategies that we deployed in the real world for optimizing ML fleet performance with the help of this metric. We show how MPG helps us identify compiler optimizations to generate more efficient code for specific hardware configurations, and improve scheduling algorithms to better utilize heterogeneous resources within the fleet, along with a few other strategies. Even after optimizations are implemented in the fleet, the MPG metric can be used to validate and track fleet-wide improvements.

These contributions collectively provide a basis for understanding and optimizing large-scale ML infrastructure, thereby paving the way for more efficient, sustainable, and cost-effective deployment of ML workloads across diverse computing environments.

\section{Background}\label{sec:background}
We introduce the concept of an ML fleet, a specialized form of computing infrastructure designed specifically for ML workloads at scale. To help set the stage for an ML fleet, we compare ML fleets with two related large-scale computing paradigms: traditional warehouse-scale computers (WSC) \cite{barroso2009wsc} and high-performance computing (HPC) supercomputers \cite{asanovic2009berkeleyview}. An ML fleet can be considered an evolution of a WSC, adapted for high-performance ML workloads. 

Table~\ref{tab:comparison} summarizes the key features of each system type, highlighting their distinct characteristics. We examine these features to better understand how the unique demands of ML workloads have shaped the design and optimization strategies of ML fleets, setting them apart from their predecessors in large-scale computing.
\subsection{Workloads}
WSCs~\cite{barroso2009wsc} traditionally focus on Internet-scale services, which serve web content, process user queries, and handle data storage and retrieval for billions of users. WSCs must manage diverse, often latency-sensitive tasks such as search indexing \cite{brin1998anatomy}, social media feeds \cite{nishtala2013memcache}, and transaction processing services. The workloads are characterized by their burstiness and high concurrency, with millions of small, independent tasks that require fast response times.

High-performance computing, on the other hand, has focused on simulation models for research, such as climate modeling \cite{menemenlis2005nasa}, and molecular dynamics for drug discovery \cite{schulz2009scaling}. These workloads involve complex, tightly-coupled computations that require both significant processing power and high-bandwidth, low-latency communication between nodes \cite{stegailov2019angara}. They operate on large datasets and require sustained performance over long periods of execution.

ML fleets combine the scale of WSC workloads with the computational intensity of HPC workloads. They consist of workloads which are focused on running deep neural networks \cite{alexnet2012}, such as recommendation models \citep{naumov2019deeplearningrecommendationmodel,zhao2022understanding} for e-commerce and social media, computer vision tasks for autonomous vehicles \cite{janai2020computer} and medical imaging \cite{esteva2021deep}, and natural language processing (NLP) applications like large language models for chatbots and translation services~\cite{zhao2021understanding}.  ML workloads involve heavy matrix-matrix \cite{fatahalian2004understanding} or matrix-vector \cite{bell2008efficient} operations and parallel processing that often require high computational throughput and memory bandwidth. ML workloads also tend to be more communication-bound rather than compute-bound \cite{li2014communication}, especially since large models are often sharded across many chips and require large datasets to be fed as input. ML fleets also face another key challenge, which is rapidly fluctuating user demand and huge tectonic shifts in ML model architectures, as discussed in \autoref{sec:intro}. As a result, the designers of an ML fleet need to trade off specialization with fungibility to adapt quickly while still delivering peak performance. 
\subsection{Hardware}
The hardware composition of these systems also reflects the demands of its users. WSCs usually rely on general-purpose commodity CPUs~\cite{barroso2009wsc}, balancing diverse computational needs for web services, and scaling massively to accommodate varying user demands \cite{ranganathan2024twentyfive}. HPC supercomputers, designed for complex scientific calculations, utilize specialized vector processors \cite{odajima2020fujitsu} and accelerators to achieve extreme performance for tightly-coupled computations. An ML fleet must utilize hardware that can provide both the scalability of a WSC and the performance intensity of a HPC supercomputer. They are mostly comprised of accelerators such as GPUs, TPUs, and/or other ML application-specific integrated circuits (ASICs) that are optimized for the compute-intensive parallel processing demands of machine learning tasks. They also make significant use of general-purpose chips like CPUs for hosting or scheduling tasks.

Memory and storage architectures are also highly tailored to their specific workloads' needs. ML fleets typically employ high-throughput, local memory systems coupled with ML-optimized SSD storage, which facilitates fast data access and parallelism for iterative learning algorithms\cite{kumar2021exploring}.  WSCs, on the other hand, utilize distributed, commodity-based memory and storage systems, prioritizing cost-effectiveness and redundancy for handling diverse, large-scale web services \cite{barroso2009wsc}. HPC supercomputers feature low-latency, parallel-optimized memory architectures and parallel file systems, enabling efficient processing of massive scientific datasets. These distinctions in memory and storage design influence each system's data handling capabilities: WSCs ensure robust, scalable data management for varied web applications; HPC systems focus on minimizing latency for complex, data-intensive scientific computations; and ML fleets optimize for repeated, high-throughput access to training data and weight updates.  
\subsection{System Efficiency}

Network infrastructure and job scheduling in ML fleets, WSCs, and HPC supercomputers are also optimized for their respective workloads. WSCs employ distribution-optimized networks to handle diverse, geographically dispersed web traffic, using cloud-based or custom schedulers \cite{burns2016borg} designed for varied, often short-lived tasks. HPC supercomputers, however, feature ultra-low latency networks crucial for tightly-coupled parallel computations, alongside specialized job schedulers like SLURM\cite{yoo2003slurm} that manage complex, long-running scientific workloads. ML fleets prioritize high-bandwidth networks \cite{zu2024resiliency} to facilitate rapid data movement for distributed training, coupled with job schedulers like Borg \cite{verma2015borg}, Kubernetes~\cite{rensin2015kubernetes} and MAST\cite{choudhury2024mast} that efficiently manage GPU/TPU resources. These differences in network and scheduling approaches directly impact each system's ability to handle its target applications.

Efficiency considerations and workloads vary across ML fleets, WSCs, and HPC supercomputers. ML fleets exhibit workload dependent energy efficiency, with high initial costs due to model training, but improved efficiency with tasks like bulk inference. WSCs prioritize high energy efficiency at scale and low per-unit costs, despite high aggregate expenses, handling diverse web services \cite{spanner} and cloud computing workloads. Meanwhile, HPC supercomputers focus on maximum performance for tightly-coupled scientific simulations, often at the expense of energy efficiency\cite{kamil2008power}. 

These distinctions in efficiency and workload optimization reflect each system's primary purpose: ML fleets are designed for flexible, scalable AI computation; WSCs for cost-effective, large-scale web services; and HPC systems for pushing the boundaries of computational performance in scientific research. The trade-offs between cost, energy efficiency, and performance in each system are carefully balanced to best serve their intended applications. In particular, if we wish to measure the performance of the ML fleet and identify opportunities for further optimization, we must study its characteristics across all levels of the system stack.

\section{Anatomy of an ML Fleet}\label{sec:fleet}

\begin{figure}[t]
    \centering
    \includegraphics[width=\columnwidth]{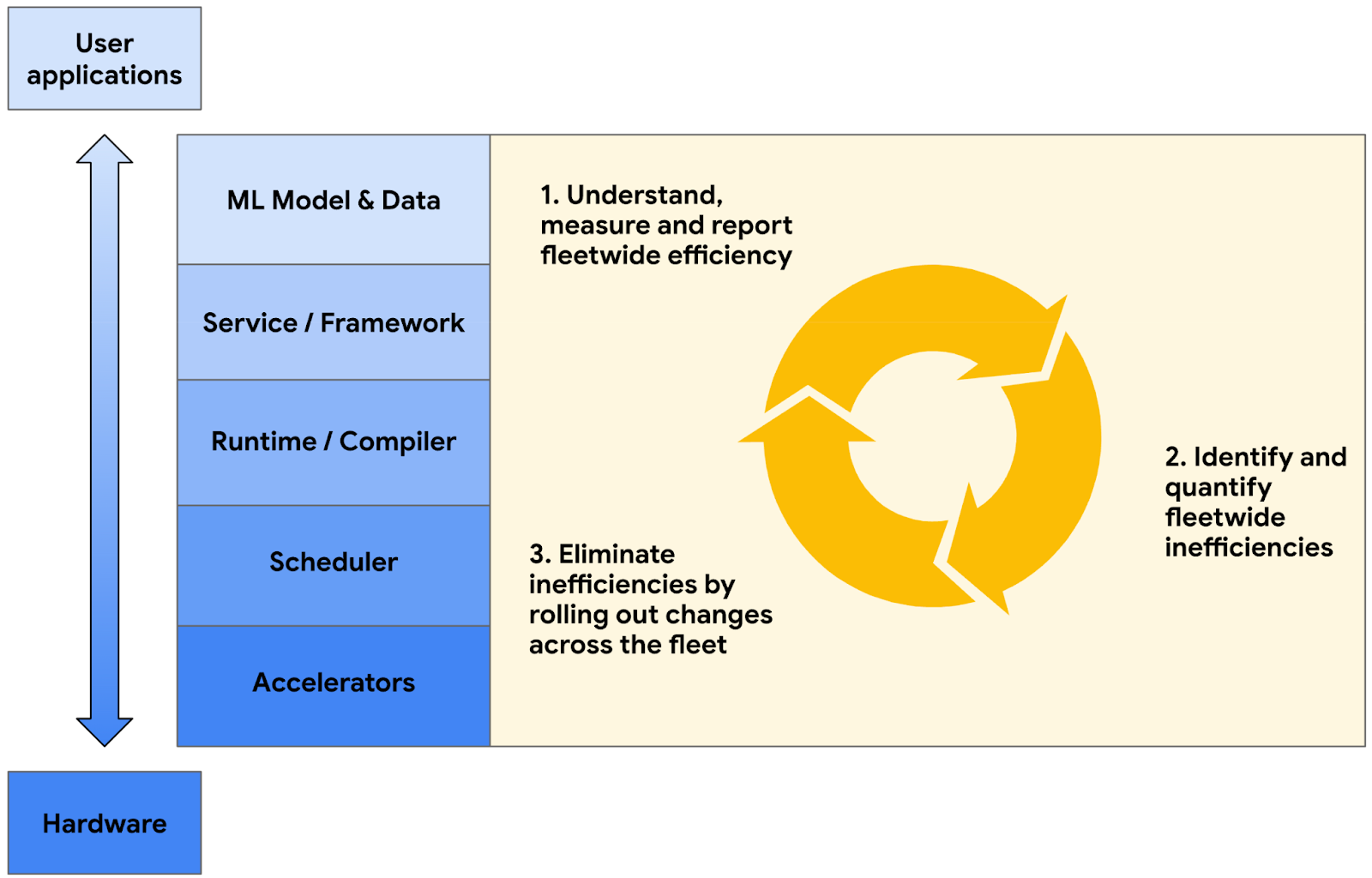}
    \caption{The ML fleet system stack of a production system at Google. The multi-layered architecture of a fleet is complex; each layer is a critical component in the ML system, with interactions between layers affecting overall performance and efficiency. Segmenting the fleet based on these layers provides actionable metrics which can be used to improve performance. }
    \label{fig:ml_stack}
\end{figure}
We begin by dissecting Google's user-facing production ML fleet, analyzing its components from hardware foundation to application level. \autoref{fig:ml_stack} shows the layers comprising the system stack that mediate user access. We supplement our discussion with data from a snapshot of Google's TPU fleet for internal workloads, providing concrete examples of challenges these systems face. %
\subsection{Accelerators}
ML fleets are distinguished from other types of large-scale systems by their accelerator-centric architecture. The ML computing landscape is dominated by domain-specific hardware, such as GPUs and other ASICs. In order to tailor to the vector and matrix intensive operations that underpin ML workloads, there has been a Cambrian explosion of ML hardware accelerators \cite{hennessy2019anewgoldenage}, with new accelerators being deployed at an unprecedented rate compared to traditional WSC fleets \cite{jouppi2021ten}. \autoref{fig:five-years} illustrates this dynamism, showing dramatic shifts in our ML fleet's hardware makeup for internal workloads over just a few years.

ML fleets typically incorporate a diverse array of hardware including CPUs, GPUs, TPUs, and other accelerators, each fulfilling specific roles. For example, CPUs may be responsible for scheduling, GPUs for training tasks, and edge accelerators \cite{yazdanbakhsh2021edge} for deployment and serving. The challenge lies in effectively orchestrating these heterogeneous accelerators to maximize their individual strengths---a complexity rarely encountered in general compute fleets.

Moreover, the heterogeneity extends beyond just accelerator type. Even within a single class of hardware accelerators, there are many different versions of the hardware, adding another layer of complexity to fleet management. Each hardware generation introduces unique features that require significant optimizations to extract peak ML workload efficiency. One notable example is the integration of the SparseCore (SC) in TPUv4 \cite{jouppi2023tpuv4opticallyreconfigurable}, which was designed to significantly boost performance for embedding-heavy models. Subsequent large-embedding model teams would likely then consider the hardware specifications of the SparseCore when designing their embedding configurations. Model design configurations such as embedding dimension, vocabulary size, valence, and others might also be co-designed to optimize performance on the hardware platform. This shows how hardware-software co-design is becoming increasingly important in improving the efficiency of these diverse accelerators, forming a symbiotic relationship where the computational needs of future workloads affect the next generation of hardware, and the hardware capabilities inform the types of workloads that the ML fleet is best equipped to handle \cite{shi2020learned}.

\subsection{Scheduler}\label{sec:scheduler}
\begin{figure}[t!]
    \centering
    \includegraphics[width=3in]{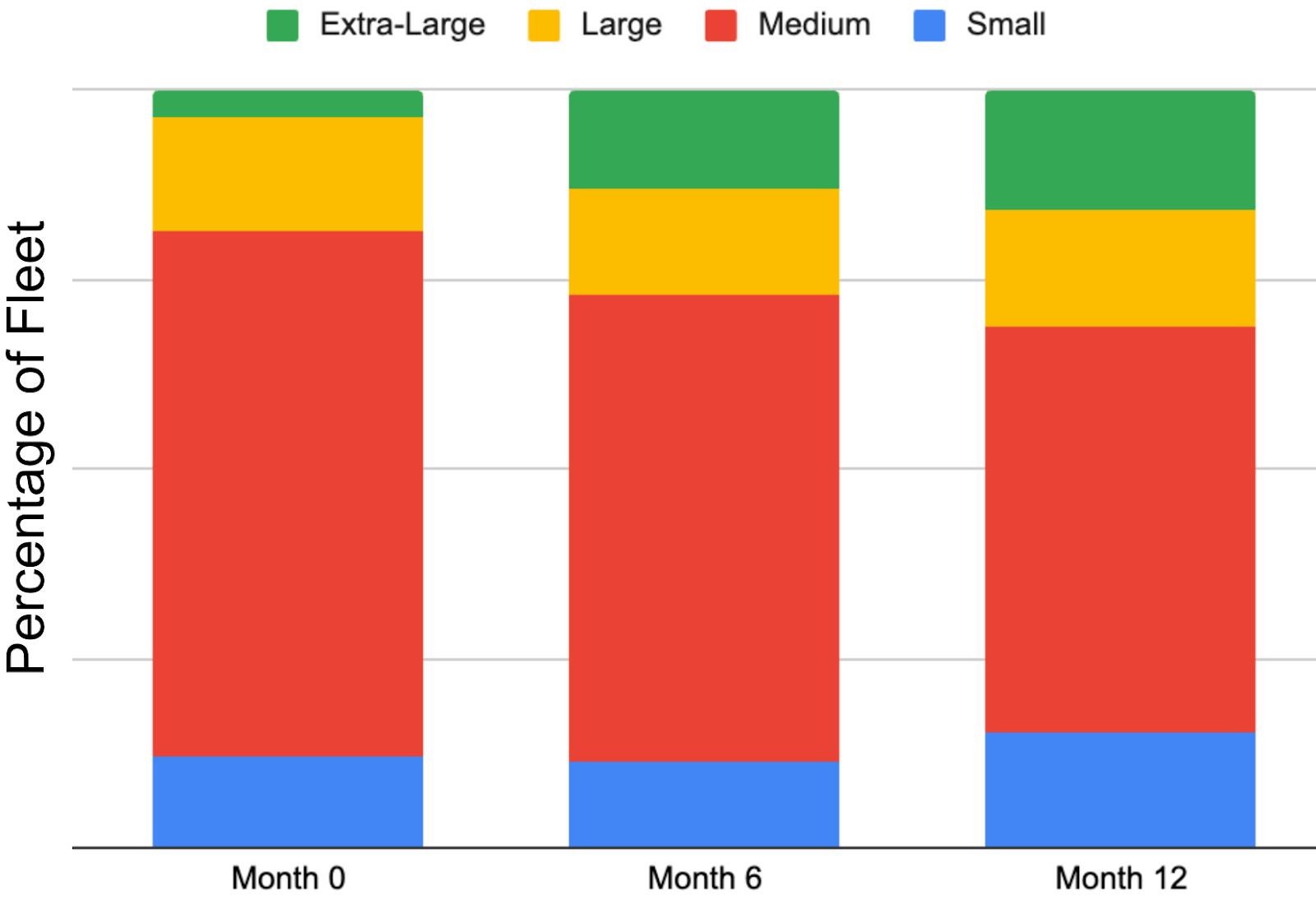}
    \caption{A sample breakdown of Google's ML fleet for internal workloads, segmenting on workload topology size (the number of accelerators requested by a given job). Progressive snapshots over the course of one year illustrate the ML fleet's growing share of jobs using an "extra-large" number of accelerators. This demonstrates how an ML fleet scheduler must be able to adapt to changing conditions, as the evolution of job sizes and topologies in response to shifting ML workloads presents unique challenges for the entire fleet.}
    \label{fig:job_size}
\end{figure}
The scheduler directly manages the hardware in a fleet by coordinating the allocation of resources. For the case study presented in this paper, it coordinates TPU allocations for Google's internal-facing ML workloads. There are two interconnected challenges that a scheduler must address when allocating hardware for an ML fleet: (1)~optimizing performance across various hardware types, and (2)~balancing utilization with stability and fault tolerance.

\autoref{fig:job_size} illustrates these challenges. It shows the allocation of workloads in Google's internal-facing ML fleet with different chip requirements over time, categorized into sizes based on the total number of TPU chips in the required topology. In this categorization, workloads with size "small" refer to jobs that request a single TPU or a handful of TPUs, while workloads with size "extra-large" refer to jobs that request the largest number of TPUs (often requiring multiple pods, as described in \citet{kumar2021exploring}). \autoref{fig:job_size} demonstrates that over the course of just one year, the allocation distribution can shift dramatically, reflecting the changing nature of ML workloads in the fleet. As large-scale ML models become more prevalent in an ML fleet, an increasing number of workloads will require correspondingly larger meshes of connected accelerators.

Optimizing the scheduling of jobs while meeting these resource requirements is difficult because it presents an NP-hard bin-packing problem. Each workload may specify a different accelerator type, chip topology, and location requirement and needs to be scheduled according to fleet constraints in a way that reduces overall fragmentation of the fleet. Since workloads are constantly being started and completed, the machine availability of the fleet is constantly changing, requiring a robust defragmentation algorithm. In addition, latency requirements may require accelerators for a workload to be grouped together near certain locations or data cells, adding another constraint to the scheduling optimization problem. 

The utilization of fleet resources must also be balanced with stability and fault tolerance. For example, to reduce disruptions, some machines may intentionally remain underutilized so that higher priority jobs may be more easily scheduled when needed. While high utilization is desirable for cost-efficiency, pushing hardware to its limits can lead to thermal issues, increased failure rates, and unpredictable performance. In large-scale ML fleets, hardware failures are inevitable, and the scheduler must be robust enough to handle these failures gracefully, redistributing workloads and ensuring job continuity without significant performance degradation.

\begin{figure}[t!]
    \centering
    \includegraphics[width=3in]{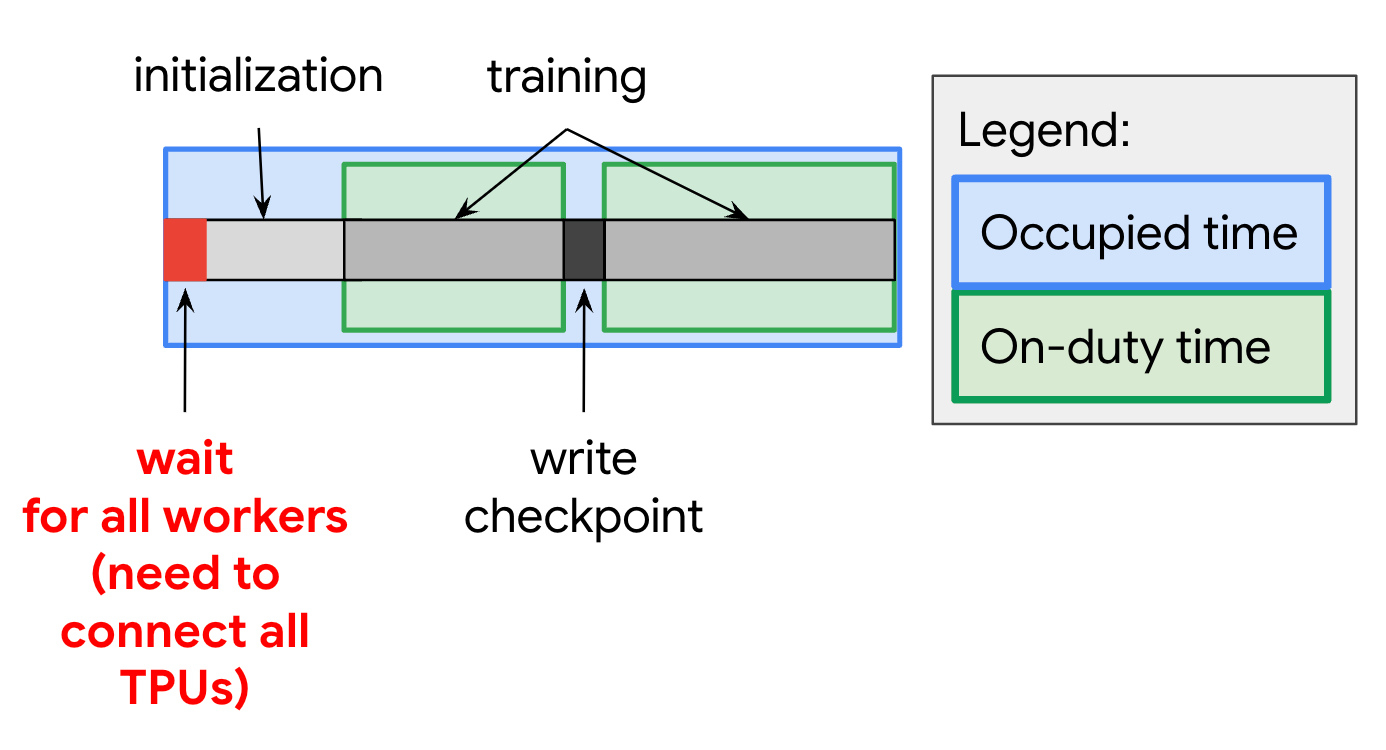}
    \caption{An ML workload requires all requested TPUs to be allocated before the task can start. In this example of a training workload, forward progress is saved via checkpoints. Delays during workload initialization and checkpoint writing, which are part of the Runtime and Framework layers, can reduce overall system efficiency.}
    \label{fig:life_of_mlapp}
\end{figure}

\subsection{Runtime/Compiler}

The runtime and compiler layers form an important component in the ML fleet system stack. They are responsible for bridging the gap between high-level ML models and the underlying hardware accelerators. The runtime layer focuses on the execution environment of ML programs. It handles important tasks such as program setup, data feeding, result management, and checkpoint creation, as illustrated in \autoref{fig:life_of_mlapp}. Depending on the system design, it either triggers just-in-time compilation of user-written code into accelerator-specific instructions or invokes pre-compiled operation kernels from vendor-specific libraries. The runtime layer can also manage the distribution strategy of code execution, as with notable runtimes like Pathways \cite{barham2022pathways}.

\begin{figure}[t!]
    \centering
    \includegraphics[width=\columnwidth]{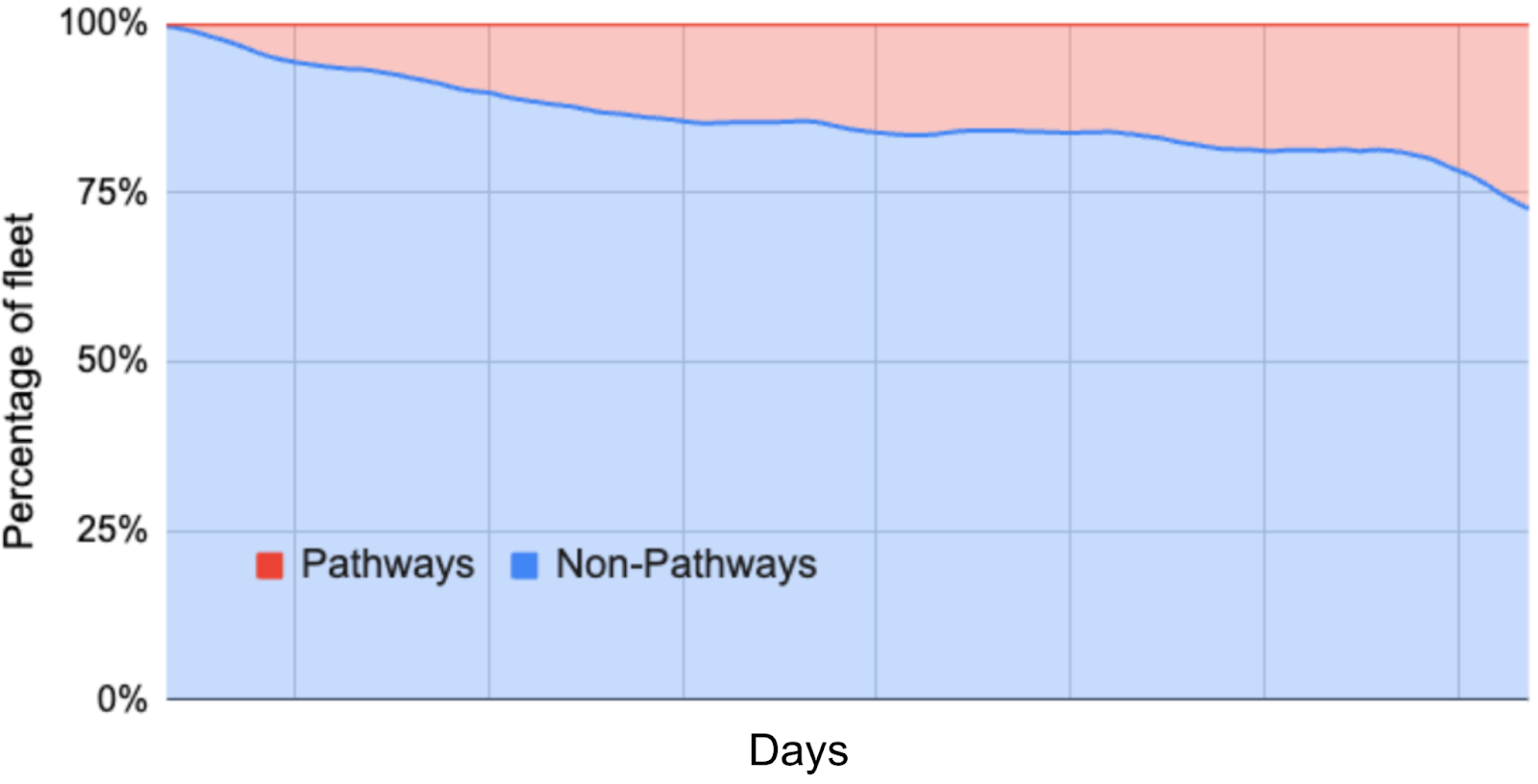}
    \caption{The prevalence of fleet-wide workloads using the Pathways runtime over a sample of one year, illustrating the rapid shift of fleet-wide runtimes to accommodate changing workloads. Pathways adoption has increased rapidly, as it provides better support for distributed execution and data processing.}
    \label{fig:pathways}
\end{figure}

\autoref{fig:pathways} shows the growth of Pathways-based workloads in our production fleet. It highlights the demand for runtimes that support efficient distributed execution for ML workloads. It also emphasizes the rapidly shifting distribution of workload runtimes in a fleet.

The compiler layer, working with the runtime, transforms high-level ML model code into executable code optimized for specific accelerators. It operates on graph intermediate representations, applying both platform-independent and platform-dependent optimizations. The output is a program tailored to the target accelerator, such as a specific version of a TPU. Domain-specific compilers, like XLA (Accelerated Linear Algebra) \cite{xla}, have significantly improved the performance of ML workloads. For instance, in MLPerf BERT benchmarks \cite{mattson2020mlperf}, XLA demonstrated a remarkable 7$\times$ performance boost and 5$\times$ batch size improvement \cite{kumar2021exploring} over previous records, emphasizing the potential of specialized compilation techniques. We note that there are many types of accelerators, some of which do not require an explicit compiler for code generation. 

Compiler optimization in ML fleets faces unique challenges due to the rapid evolution of hardware accelerators, requiring frequent updating of optimization strategies to leverage the specific features of each new hardware generation. Moreover, the impact of optimizations can be difficult to generalize, as an optimization that improves one workload may degrade another due to differences in computation or communication patterns. This emphasizes the need for a balanced approach to optimization, considering both platform-independent techniques for flexibility and platform-specific optimizations for maximum performance.

\subsection{Framework}

The framework layer sits on top of the runtime/compiler. It is the interface between ML practitioners and the underlying complex hardware and software infrastructure. This layer encompasses various ML frameworks and libraries, such as TensorFlow \cite{abadi2016tensorflow}, JAX \cite{frostig2018compiling}, and PyTorch \cite{paszke2019pytorch}, each offering unique features and optimizations. 

The framework layer provides high-level abstractions and APIs that allow developers to build and deploy ML models efficiently. These frameworks are responsible for translating user-written code into representations that can be understood and optimized by lower-level layers such as compilers and runtimes. This translation process bridges the gap between user intent and system execution.

\begin{figure}[t!]
    \centering
    \includegraphics[width=.75\linewidth]{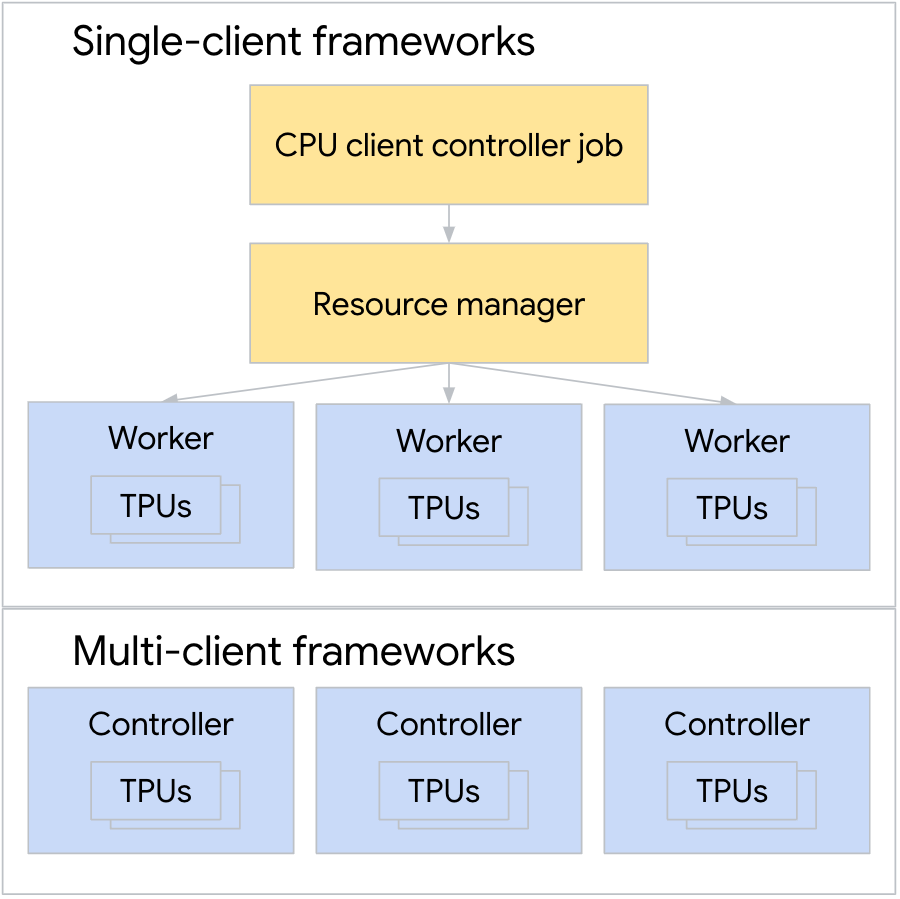}
    \caption{
Comparing single-client frameworks with multi-client frameworks. 
    }
    \label{fig:multi_single_frameworks}
\end{figure}

One of the key responsibilities of ML frameworks is defining the structure of distributed ML applications. For example, TensorFlow's Distribution Strategy \cite{abadi2016tensorflowlargescalemachinelearning} provides a framework for distributing training across multiple devices or machines. These can have single-client or multi-client architectures, depending on workload needs, as shown in \autoref{fig:multi_single_frameworks}. These frameworks must also map ML primitives to hardware-specific designs to achieve optimal performance. This is important for specialized hardware like TPUs, which are designed for bulk-synchronous training. Frameworks like JAX are more targeted towards ML workloads, with features that facilitate ease of interpretability when analyzing ML performance, such as high-level tracing for just-in-time compilation. In the ML fleet, JAX usage has increased over time, most likely due to these features and the emergence of more ML-heavy workloads \cite{frostig2018compiling}.

Also, ML frameworks often provide auxiliary services to improve efficiency of the entire ML fleet. For instance, TensorFlow's \texttt{tf.data} \cite{murray2021tfdata} service optimizes the performance of the data pipeline. These features, while abstracted from the user, can impact the overall system efficiency, as shown in \autoref{fig:life_of_mlapp}. Underneath these high-level frameworks lies a foundation of general-purpose libraries and datacenter services. Frameworks like TensorFlow utilize libraries such as gRPC \cite{grpc}, protobuf \cite{protobuf}, and tcmalloc \cite{tcmalloc} for various low-level operations, and interface with datacenter services for storage (e.g., Colossus \cite{ghemawat2003gfs} \cite{colossus}) and monitoring (e.g., Monarch \cite{adams2020monarch}).

As the primary point of interaction for users, the framework layer serves as a key bridge in the ML system stack, as it not only abstracts underlying complexities but also plays an important role in determining the overall efficiency and capabilities of the fleet. Frameworks must balance the need for user-friendly APIs with the need to leverage underlying hardware-specific optimizations, while also managing the complexities of distributed computing, data pipeline optimization, and integration with lower-level services.

\subsection{ML Model \& Data}

To characterize the ML Fleet at the highest level of the stack, we  want to know: What types of workloads are we spending most of our compute cycles on? This is a critical question because it drives nearly every design decision we make for the ML Fleet at every level of the stack, from the hardware (how many training vs. inference vs. other chips) to the software (JAX vs. other frameworks, runtime distribution strategies, and compiler optimizations). Workload heterogeneity analysis is useful for understanding what kind of models are prevalent in the production fleet, especially since different workloads stress the hardware in different ways. This understanding can drive decisions about which accelerators to deploy and how many, or which compiler optimizations to carry out.

In practice, we observe that the model and data layer of the ML fleet stack are the most affected by fluctuating user demands. User requirements such as the model architecture, size of the training dataset, or even use of different numerical formats in the training model can impact the efficiency of the job, which can have a cascading effect on the efficiency of the overall ML fleet.

While ML workloads share some computational patterns, particularly in their use of matrix operations and data-intensive processing, the specific architectures and resource requirements can vary significantly.  As new model architectures and learning tasks emerge, they prompt rapid shifts in workload composition, leading to fluctuations in resource demands across various model types. 

In a production ML fleet, there are varying proportions of workloads dedicated to each phase of the ML model life cycle; training, bulk inference, and real-time serving. Thus, the fleet must be flexible enough to handle the requirements of each of these phases; for example, training workloads may be compute intensive while real-time serving workloads may focus on minimizing latency.

\section{ML Productivity Goodput}\label{sec:goodput}

The optimization of ML fleet efficiency is a complex, cyclical challenge, as illustrated in \autoref{fig:ml_stack}. The first challenge is measuring, understanding, and reporting fleetwide efficiency, establishing a baseline for current performance. Second, we must identify and quantify fleet-wide inefficiencies, pinpointing areas that require improvement. Third, we must eliminate these inefficiencies by implementing changes across the fleet, which in turn leads back to the first stage as we measure the impact of these changes. This cycle ensures ongoing optimization and adaptation to the ever-evolving landscape of ML workloads and hardware. To keep pace with this lifecycle, we require a metric that not only quantifies current performance but also guides optimization efforts across the fleet. %

\begin{figure}[t!]
    \centering
    \includegraphics[width=\columnwidth]{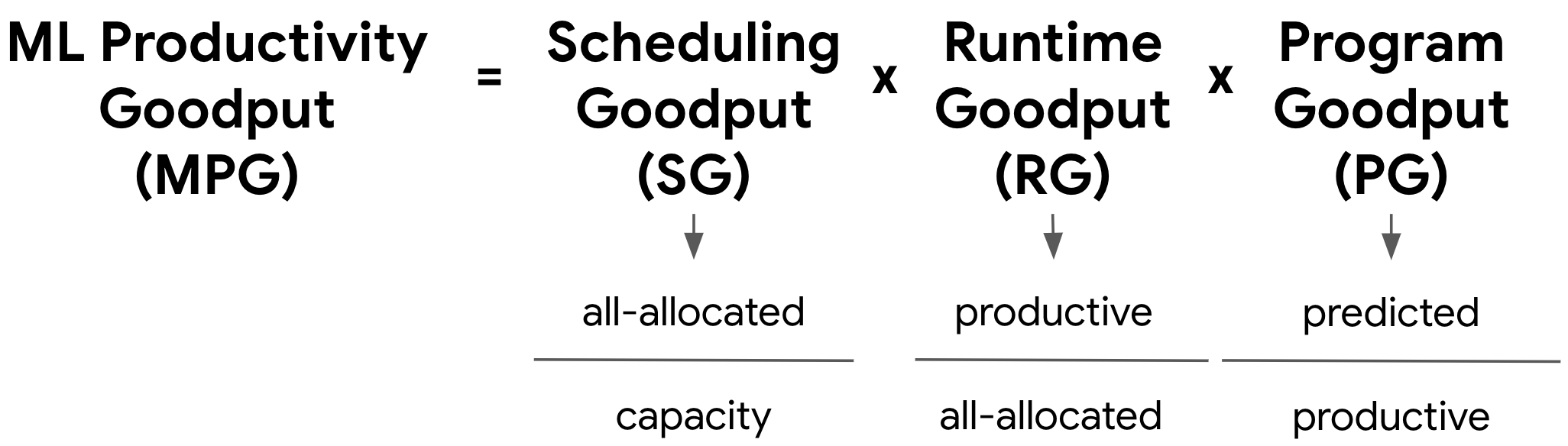}
    \caption{ML Productivity Goodput (MPG) and its components.}
    \label{fig:mpg}
\end{figure}

In this section, we present an in-depth discussion of MPG, a new metric for quantifying ML fleet efficiency. We refer to this as the iron law of performance for ML fleets, drawing a corollary to the iron law of processor performance~\cite{emer1984ironlaw}. MPG, defined in \autoref{fig:mpg}, is a means for measuring efficiency gains and guiding exploration of optimization strategies across various fleet components.

\subsection{Metric Features}
Ideally, the MPG metric must be a clearly defined and accurate measure of forward progress; improvements in the metric must also reflect real improvements in the efficiency of the fleet. This metric must be capable of overcoming two significant challenges.

\begin{enumerate}
\item \textbf{It must capture the dynamic nature of ML fleets}: The fleet is constantly fluctuating due to variables such as changes in workload composition, updates to the code stack, and evolving hardware. To effectively improve efficiency, we must ensure that any change in the metric is explainable despite these fluctuating variables.
\item \textbf{It must explain the trade-offs between individual and aggregate efficiency}. At a fleetwide scale, jobs must be scheduled in concert with one another to ensure maximum aggregate efficiency of the fleet. However, individual jobs may have certain service-level requirements, meaning that this metric must be decomposable based on workload characteristics.
\end{enumerate}

\begin{figure}[t]
    \centering
    \includegraphics[width=\columnwidth]{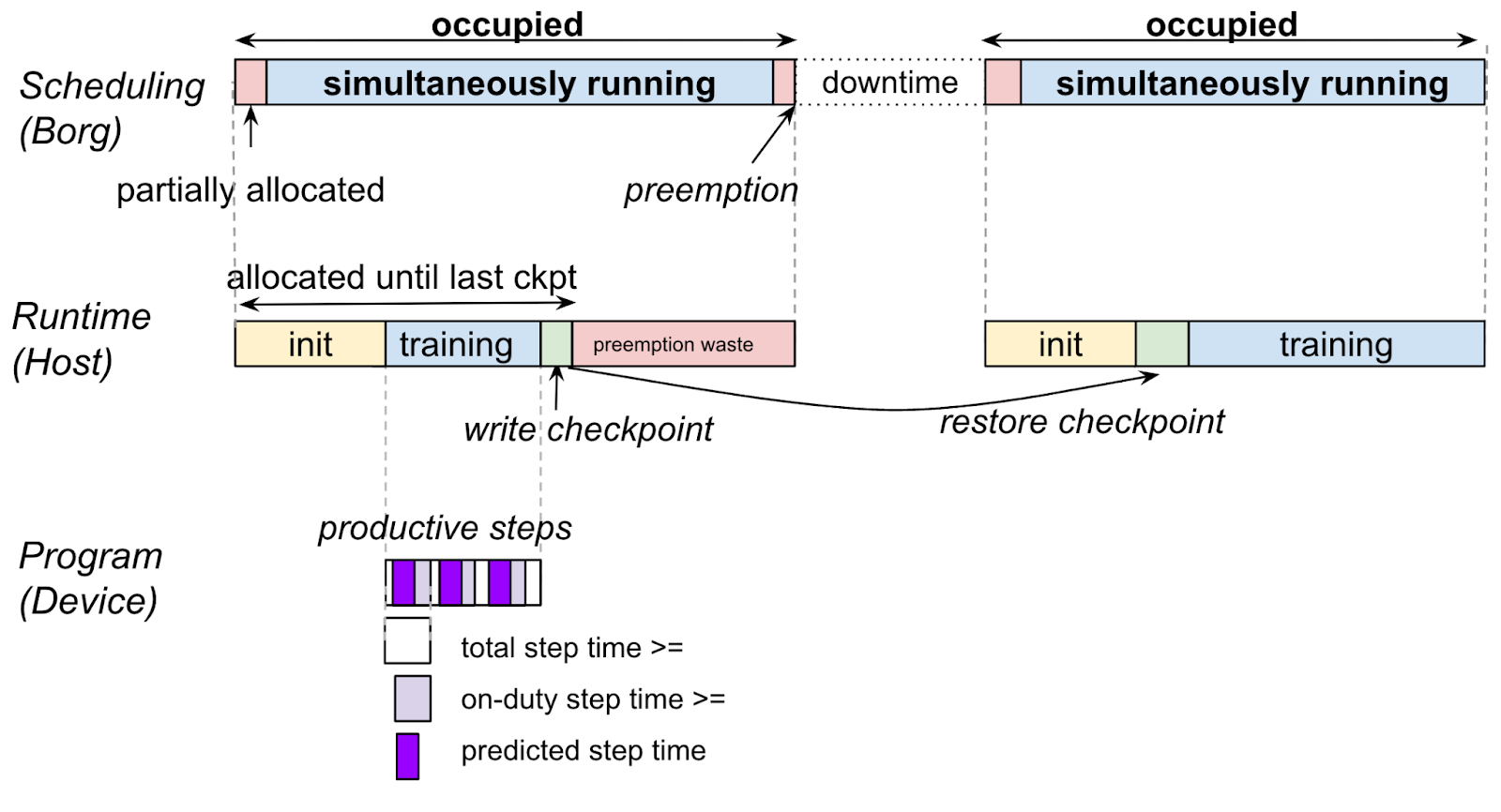}
    \caption{Breakdown of a ML workload using \mpg. 
    }
    \label{fig:mpg_breakdown}
\end{figure}

\subsection{A New Approach: ML Productivity Goodput}

\mpg (MPG) is designed to address the myriad challenges discussed in Section~\ref{sec:fleet}, as well as to overcome the limitations of existing approaches. Just as the Iron Law of Processor Performance~\cite{emer1984ironlaw} breaks down CPU performance into $\frac{instructions}{program}$$\times$ $\frac{cycles}{instruction}$$\times$$\frac{time}{cycle}$, the MPG metric decomposes ML fleet efficiency into scheduling, runtime, and program components (see \autoref{fig:mpg}).

This multi-layered structure, as shown in \autoref{fig:mpg_breakdown}, offers several advantages over traditional metrics. First, it allows precise identification of performance bottlenecks or improvements at specific layers of the stack, facilitates a more granular analysis of efficiency trends over time, and mitigates the risk of misleading interpretations that can arise from aggregated data. Second, by decoupling these submetrics, we enable more targeted optimization efforts and gain deeper insights into the complex interactions within the ML fleet. Finally, this approach not only enhances our ability to measure current performance but also provides a framework for guiding improvements, discussed in \autoref{sec:improvements}.  

\textbf{Scheduling Goodput:}
\emph{How often does an application have all necessary resources to make progress?} 

\begin{figure}[t!]
    \centering
    \includegraphics[width=\columnwidth]{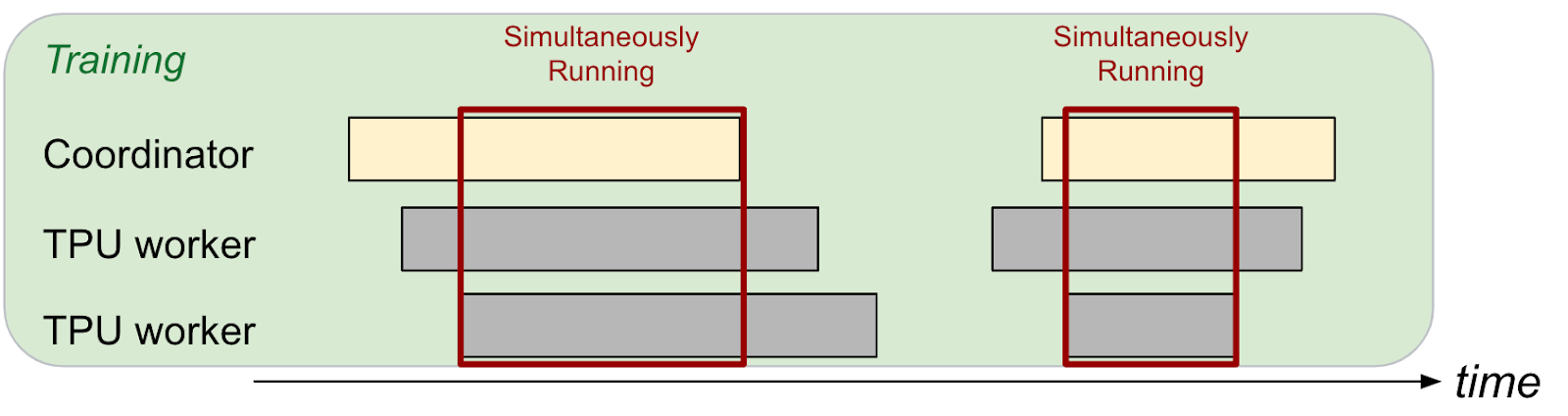}
    \caption{The scheduling goodput for training workloads measures the percentage of time when all of the TPU workers are available to work at the same time. In other words, it measures the portion of time that all of the necessary resources are available to make progress. 
}
    \label{fig:scheduling}
\end{figure}

Scheduling Goodput (SG) quantifies the efficiency of resource allocation in an ML fleet. It measures the fraction of time that an application has all the required resources simultaneously available to make progress. This metric can be lower than traditional Occupancy metrics, particularly in distributed, bulk-synchronous applications where all required chips must be available concurrently. The numerator of SG is calculated as the simultaneous uptime of all tasks in a distributed ML application that must be connected to make synchronous progress, as shown in \autoref{fig:scheduling}. This is referred to as ``allocated chip-time'' or ``all-allocated'' time. The denominator is fleet capacity, expressed as chip-time. This provides a full view of how effectively the scheduling layer is using the fleet's resources. 

Scheduling Goodput offers insights into potential inefficiencies in resource allocation, such as fragmentation of available resources, delays in coordinating multiple chips for distributed applications, and mismatches between application requirements and available resources. By optimizing SG, we can improve the overall efficiency of resource utilization in the ML fleet, ensuring that applications have the necessary resources to make consistent progress.

\textbf{Runtime Goodput:}
\emph{Of the time that an application has all necessary resources, how often is it making progress?} 

Runtime Goodput (RG) measures the efficiency of the orchestration layers in managing the execution of ML applications once resources are allocated. This metric focuses on the actual productive time of an application, accounting for various overheads in the runtime environment. The orchestration layer is responsible for critical tasks such as initializing chips, connecting them into slices for bulk-synchronous progress, loading and compiling programs, feeding data to these programs, and ensuring that training progress is regularly saved through checkpoints. The numerator of RG is the productive chip-time of the application's progress that has been saved in checkpoints; work done between the last checkpoint and failure (or preemption) doesn't count as "productive" time and is therefore not included in RG. The denominator of RG is the allocated chip-time defined as the numerator of SG. 

Runtime Goodput can help with identifying bottlenecks in the runtime environment, such as slow data loading, inefficient checkpointing, or suboptimal program compilation. It can guide the efforts to streamline the execution pipeline and improve the overall throughput of ML workloads.

\textbf{Program Goodput:}
\emph{Of the time that an application is making progress, how close is it to the ideal roofline?}

Program Goodput (PG) assesses the efficiency of the application code itself, measuring how effectively it utilizes the available computational resources. While a traditional roofline performance model~\cite{williams2009roofline} might seem suitable for this purpose, it falls short in capturing the true efficiency of modern ML workloads. The traditional roofline model is highly sensitive to compiler decisions, such as how ML operators are fused or rematerialized \cite{briggs1992rematerialization}, or which operands are placed in which memory space. It rewards individual ops that are close to peak utilization, but penalizes correct optimizations that result in computation graphs where the utilization may be lower, but overall execution time is shorter. 

To overcome these limitations, we use a compute-based roofline model that compares the ideal execution time of the workload against its actual execution time. The ideal predicted execution time, which is the numerator of PG, can be computed from intrinsic properties of the machine learning model being run. By analyzing the shape of the unoptimized high-level operations (HLO) graph, we can estimate how many floating point operations (FLOPs) the program would require at its theoretical peak performance. Since we are analyzing the computation graph before any compiler optimizations, this prediction is agnostic to compiler decisions. 

The denominator of PG is the actual execution time. The PG metric can thus be interpreted as a percentage reflecting how well optimized the ML program is, with a score of 100\% indicating perfect performance matching the theoretical peak.

\section{Improving Fleetwide Efficiency}\label{sec:improvements}
 \begin{figure}[t]
    \centering
    \includegraphics[width=\columnwidth]{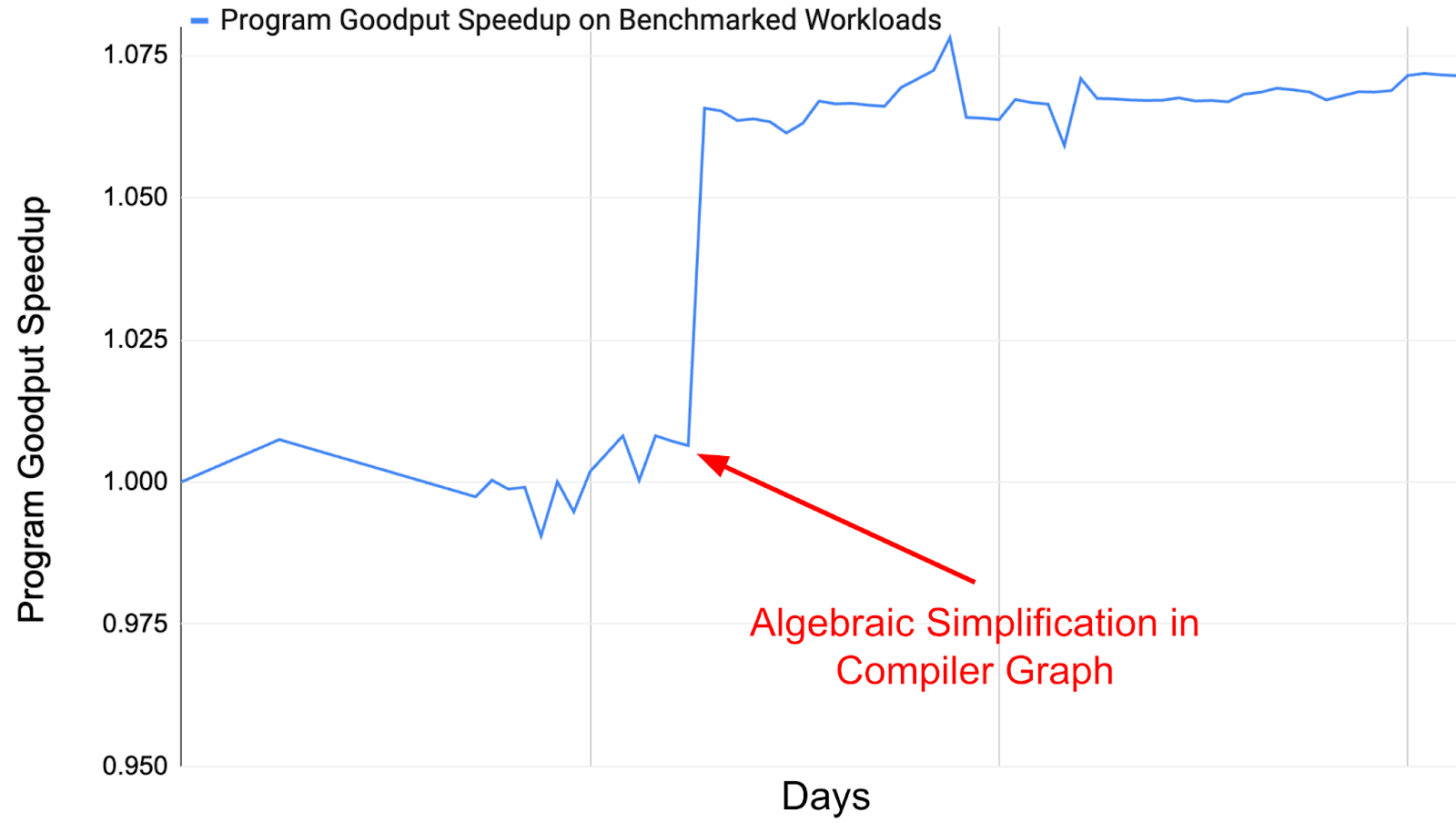}
    \caption{This figure demonstrates the effect of an XLA algebraic simplification optimization on Program Goodput (PG) across a benchmark of the top 150 fleet workloads. Looking at the PG in this way allows us to bisect which code changes improved or regressed overall fleet efficiency.}
    \label{fig:pg_cl}
\end{figure}

We show how \mpg is a robust quantifier of ML Fleet performance through optimization examples from Google's ML Fleet in production. We present a breakdown of the MPG components using segmented fleet data and demonstrate how this procedure can help identify potential optimization techniques. Additionally, we showcase the effects of deploying these optimizations and how MPG helps verify and track performance improvements.

Looking at the aggregated MPG of the fleet does not necessarily help ML practitioners identify what kinds of improvements will make the largest impact on the fleet; this is where the decomposability of the metric comes into play, as shown in \autoref{tab:improvements}. By breaking MPG into its three components; Scheduling Goodput, Runtime Goodput, and Program Goodput, we can diagnose fleetwide issues and identify the types of optimizations that would most improve fleet efficiency. Furthermore, we can segment the fleet using the characteristics described in Section~\ref{sec:fleet} in order to identify issues with specific workload types and propose model-level optimizations.

\renewcommand*{\arraystretch}{1.25}
\begin{table*}[h!]
\caption{Optimizing different components of \mpg.}
\label{tab:improvements}
\resizebox{\textwidth}{!}{%
\scriptsize
\begin{tabular}{@{}lp{2cm}p{3cm}p{3cm}p{4cm}@{}}
\toprule
\textbf{ML Fleet Stack Layer} & \textbf{Program Goodput $\times$} & \textbf{Runtime Goodput $\times$} & \textbf{Scheduling Goodput $=$} & \textbf{Workload \mpg} \\ \bottomrule
\textbf{Compiler:} \\ On-duty step time decreases &  \textbf{Increases} & Decreases if device-bound \newline Decreases if host-bound & Decreases if device-bound \newline No change if host-bound & \textbf{Increases if device-bound}  \newline No change if host-bound
 \\ \midrule
\textbf{Runtime:} \\ Off-duty time or preemption waste decreases & No change & \textbf{Increases} & Decreases & \textbf{Increases} \\ \midrule
\textbf{Scheduler:} \\ Partially-allocated time decreases & No change & No change & \textbf{Increases}  & \textbf{Increases} 
\\ \bottomrule
\end{tabular}%
}
\end{table*}
\subsection{Scheduling Goodput Optimizations}

The optimization of Scheduling Goodput (SG) can be presented as a bin packing problem, as described in \autoref{sec:scheduler}. Users launch workloads with varying TPU topology sizes, and the scheduling algorithm must determine how to best fit these workloads into the existing fleet of allocated chips. This process presents numerous challenges, primarily due to the wide variety of job sizes in the fleet, ranging from single-chip to multipod configurations \cite{kumar2021exploring}.

A significant complication in job scheduling is that it requires more than mere availability in the fleet. The topology of the available hardware must also satisfy the topology requirements of the workload, which is sometimes impossible without first pre-empting other jobs. Consequently, suboptimal scheduling can have a cascading effect on other components of \mpg.

\begin{figure}[t]
    \centering
    \includegraphics[height=1.8in]{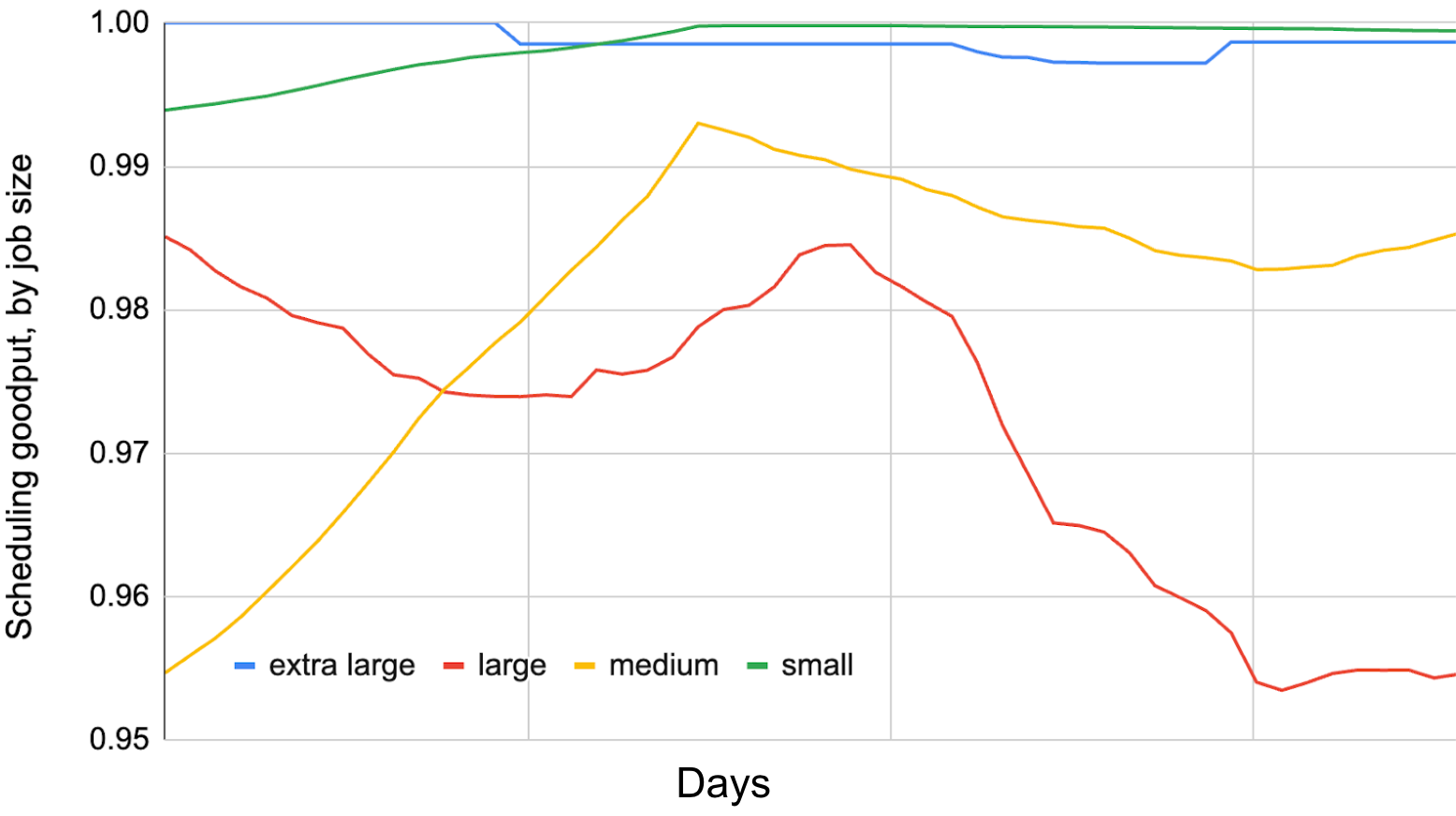}
    \caption{Scheduling goodput by job size. Extra-large and small jobs tend to have better scheduling goodput due to the scheduler's preemption algorithm.}
    \label{fig:sg_job_size}
\end{figure}

We can identify availability issues in the ML Fleet by looking at the SG for jobs with different chip allocation requirements, as shown in \autoref{fig:sg_job_size}. The data shows that the overall SG is already close to optimal, due to defragmentation techniques and scheduling optimizations. However, it is interesting to note which jobs tend to have the highest SG: extra-large jobs which require the greatest number of chips or possibly multiple TPU pods, as well as smaller jobs which require only a single chip or a few chips. 

This is likely due to the way the scheduler deals with evictions; evicting extremely large jobs would have a severe negative impact on the overall \mpg score due to their huge startup overhead. Once the extra-large job is running, it is also immensely dependent on checkpointing and data sharding, affecting the Runtime and Program Goodput components as well. In short, evicting extra-large jobs from the hardware would present a cascading series of failures, strongly incentivizing the scheduler to reduce churn for these jobs and evict medium-sized jobs instead. 

On the other hand, extremely small jobs usually do not get prematurely evicted since they are more likely to finish quickly, and if pre-empted, it is usually quicker to find topologically matching availability. With extremely small jobs, the scheduler has more flexibility to intelligently allocate the workloads to optimal compute cells in order to defragment the overall ML fleet availability. It is also important to note that for workloads of all sizes, the SG is greater than 95\% due to the particular pre-emption preferences of the scheduler. The pre-emption preferences of the scheduler can be tuned, e.g. to require a SG of greater than 95\% for medium-sized jobs, but this could reduce the SG for other segments of the fleet.

\subsection{Runtime Goodput Optimizations}
Outside of device time, host overhead and pre-emptions can be major bottlenecks for some workloads, and can be tracked by measuring Runtime Goodput. For example, training jobs usually use input pipelines to ingest and transform input data, which could be bottlenecks for certain models that ingest large amounts of data. Some solutions have been proposed to reduce host overhead, such as Plumber~\cite{kuchnik2022plumber}, a tool to find bottlenecks in ML input pipelines. 

We can improve the RG of the ML fleet with asynchronous strategies such as sharding the dataflow graph, as proposed by Pathways \cite{barham2022pathways}. The segmented analysis of RG shows that the particular workloads on Pathways tend to have higher RG scores over time, validating the benefits of Pathways for our particular ML Fleet. Also, techniques such as asynchronous checkpointing \citep{maurya2024datastates, nicolae2020deepfreeze} can reduce the time spent fetching previous model training checkpoints where the accelerators temporarily pause training and are completely idle.

Other strategies, such as ahead-of-time compilation, where programs are compiled on less expensive hardware such as CPUs and then executed on TPUs, can also improve RG. By offloading compilation to a less expensive chip and storing the results in a compilation cache, we can reduce the total runtime of more specialized accelerators. These techniques are often implemented in common ML frameworks, such as TensorFlow \cite{aotTF} and JAX \cite{aotJAX}.

\begin{figure}[t]
    \centering
    \includegraphics[width=\columnwidth]{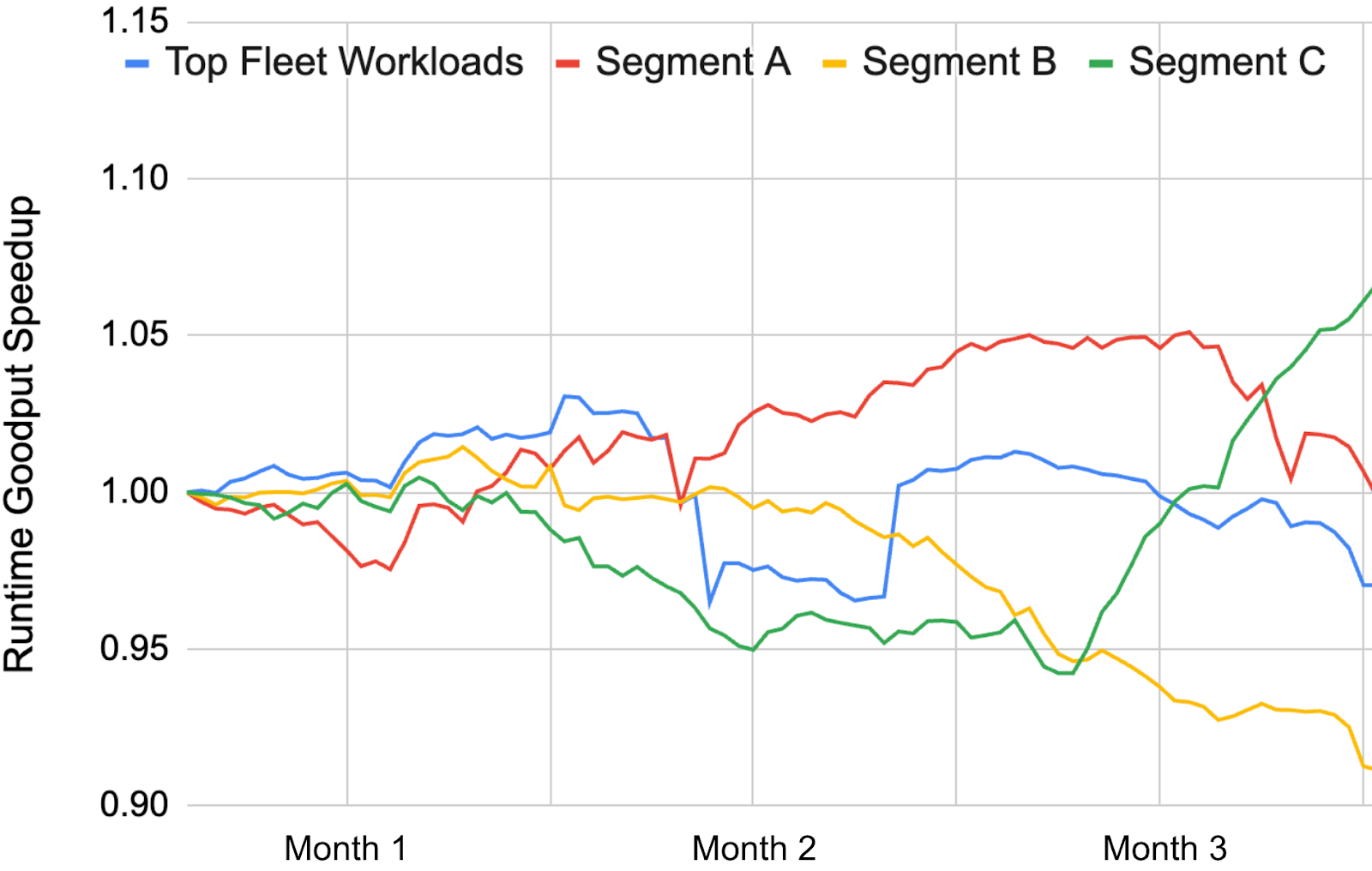}
    \caption{Runtime goodput speedups over the course of one quarter, segmented by fleet workload types. Speedup is normalized to the top N workloads in the fleet, measured at the beginning of the quarter. }
    \label{fig:segmented_runtime_goodput}
\end{figure}

To demonstrate the benefits of these framework-specific optimizations, we can examine the RG of the ML Fleet at the framework level of the system stack described in \autoref{fig:ml_stack}. This can help us understand which frameworks or runtime strategies may be better suited for which workloads.
\autoref{fig:segmented_runtime_goodput} shows RG scores from a sample of Google's ML fleet for internal workloads, segmented based on characteristics such as model architecture, product area or workload phase (training, real-time serving, or bulk inference), and compared to a baseline of top fleet workloads. Although the segments in \autoref{fig:segmented_runtime_goodput} are not explicitly identified, we demonstrate how segmenting RG based on workload characteristics (Segment~A, Segment~B, and Segment~C) can reveal trends that would otherwise be hidden by aggregate fleet metrics (represented by the "Top Fleet Workloads" segment). For example, training workloads running JAX with Pathways may tend to have a higher RG, possibly due to the fact that Pathways is single-client \cite{barham2022pathways} and therefore better optimized for training than multi-client frameworks.

\begin{figure}[h]
    \centering
    \includegraphics[width=\columnwidth]{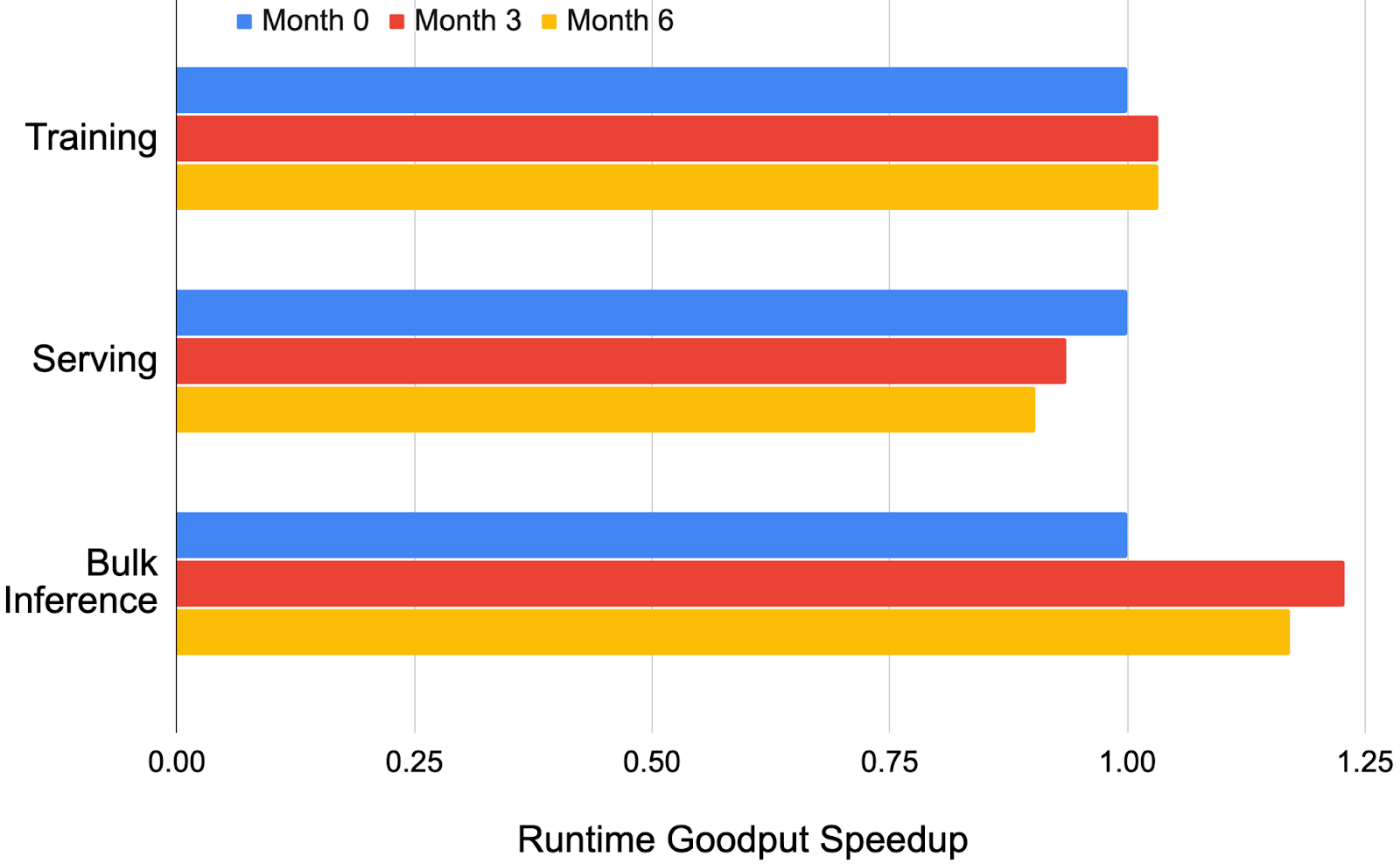}
    \caption{Runtime Goodput trends for a notional slice of a sample ML fleet over a period of six months, segmented by workload phase.}
    \label{fig:train_inf_rg}
\end{figure}

Examining the data along a different axis, training versus real-time serving versus bulk inference, can also be helpful. Using a sample from Google's ML fleet, \autoref{fig:train_inf_rg} illustrates that training workloads tend to have a higher RG than serving workloads. This is most likely due to the inherently different nature of serving and training.  Typically, training workloads have more constant computational demands, while real-time serving can fluctuate based on user demand. The slight decrease in serving RG can be attributed to transitory demands on the fleet, but it remains relatively stable in comparison to the bulk inference segment. The huge fluctuation in RG for bulk inference highlights the changing nature of production fleet demands. Previously, the bulk inference segment of the fleet was dominated by workloads running on a single core, where each chip contained a replica of the model, resulting in more easily accessible checkpoints/data and less accelerator wait time. However, as we move to larger models, the weights must be sharded across multiple chips, resulting in more expensive data reads. Additionally, the rise of expert-based models \cite{shazeer2017moe} has made bulk inference runtime much more complex to optimize, as some machines must wait for others for distillation of weight updates in a student-teacher model. This has resulted in an temporary decrease of RG for the bulk inference segment between "Month 3" and "Month 6" of \autoref{fig:train_inf_rg}. This example illustrates how analyzing disaggregated RG can allow ML fleet architects to make informed decisions about their runtime stack by pinpointing segments that may be more susceptible to shifting fleet demands.

\subsection{Program Goodput Optimizations}
We present various techniques and strategies that have been employed at \google over recent years for improving Program Goodput in our ML fleets, ranging from parallelization methods to compiler optimizations. Recall that PG measures the effective utilization of computational resources. As ML models grow in size and complexity, optimizing PG becomes increasingly important to make efficient use of hardware and reduce compute times. With PG instrumentation, we have been able to pinpoint which segments of the fleet require further optimization at the compiler or ML model level. 

 \begin{figure}[t]
    \centering
    \includegraphics[height=2in]{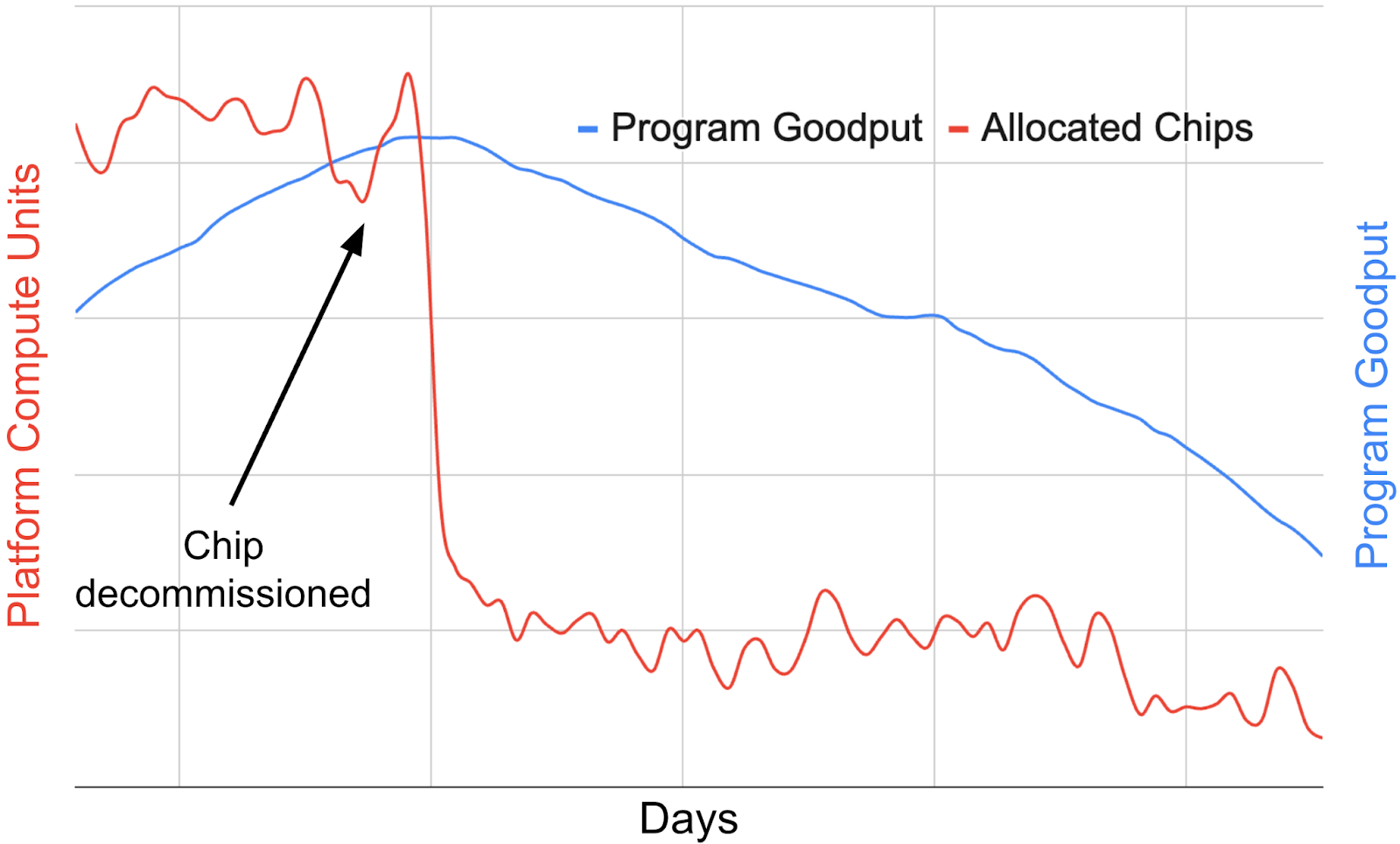}
    \caption{Tracking the Program Goodput (PG) versus allocation trends for a particular domain-specific chip in an ML fleet. Looking at the disaggregated segments of MPG can help reveal fleetwide trends and interactions between different layers in the ML fleet stack, informing future design decisions.}
    \label{fig:pg_chip_type}
\end{figure}

 \textbf{Overlapping communication and computation.}
To identify potential system optimizations that can improve fleet efficiency, we can look at the PG of workloads segmented by performance characteristics. In other words, how many of the workloads in the fleet are compute-bound versus communication-bound? By segmenting the PG in this way, it is possible for us to identify that many high-cost workloads are communication-bound. 

To address this issue at the high-level operation (HLO) level, a technique that overlaps communication with computation was developed and deployed in our production fleet (described in ~\citet{wang2022overlap}). This technique decomposes communication collectives, along with the dependent computation operations, into a sequence of finer-grained operations to hide data transfer latency so that better system utilization is achieved. This approach improved the overall system throughput by up to 1.38$\times$ and achieves 72\% FLOPS utilization on 1024 TPU chips for a large language model with 500 billion parameters~\cite{wang2022overlap}.

\textbf{Compiler autotuning.}
At the fleet level, we have also developed and deployed optimizations that improve code-generation quality and can be generalized to any workload in the fleet. XTAT~\cite{phothilimthana2021flexible} is an autotuner for production ML compilers that tunes multiple compiler stages, including tensor layouts, operator fusion decisions, tile sizes and code generation parameters. Evaluated over 150 ML training and inference models on TPUs, XTAT offers speedups over the heavily-optimized XLA compiler in the fleet.

\textbf{Example: Quantifying the impact of an XLA optimization on the TPU fleet.}
It is rare for any single optimization to have a significant impact on overall fleet-wide PG. But we can track the impact of these optimizations by looking at the change in PG for a fixed set of benchmarked workloads or segment of the production fleet over time. For example, looking at a benchmark of the top 150 most costly workloads in the fleet, \autoref{fig:pg_cl} pinpoints the effect of a code change that was submitted to the XLA compiler - in this case, an algebraic simplification in the compiler graph. The increase in PG for the benchmark of 150 workloads suggests that the positive impact of this optimization can be generalized to the fleet. 

It is also helpful to look at PG fluctuations across hardware segments of the fleet. \autoref{fig:pg_chip_type} shows a notional example where looking at segmented PG can uncover insights that would otherwise be hidden by looking at aggregate metrics. In this case, the segmented data suggests that when a new ML accelerator chip is introduced to the fleet, the workloads running on that chip may initially have a low PG, since the model / compiler code has not been fully tailored for that chip yet. As user adoption increases and accelerator-specific software optimizations are rolled out to the fleet, PG gets closer to theoretical peak efficiency. In other words, accelerator maturity tends to yield greater PG over time. As the chip nears the end of its lifecycle (represented by decreasing allocation in the fleet, and shown in \autoref{fig:pg_chip_type} by the ``Chip decommissioned'' label), the PG decreases due to lower chip usage and natural workload/compiler drift. This shows the importance of co-design across all layers of the ML fleet to make sure that both the software (compiler) and hardware (chips) are optimized for the latest workloads.

\section{Discussion}

\begin{figure}[t]
\centering
\includegraphics[width=\columnwidth]{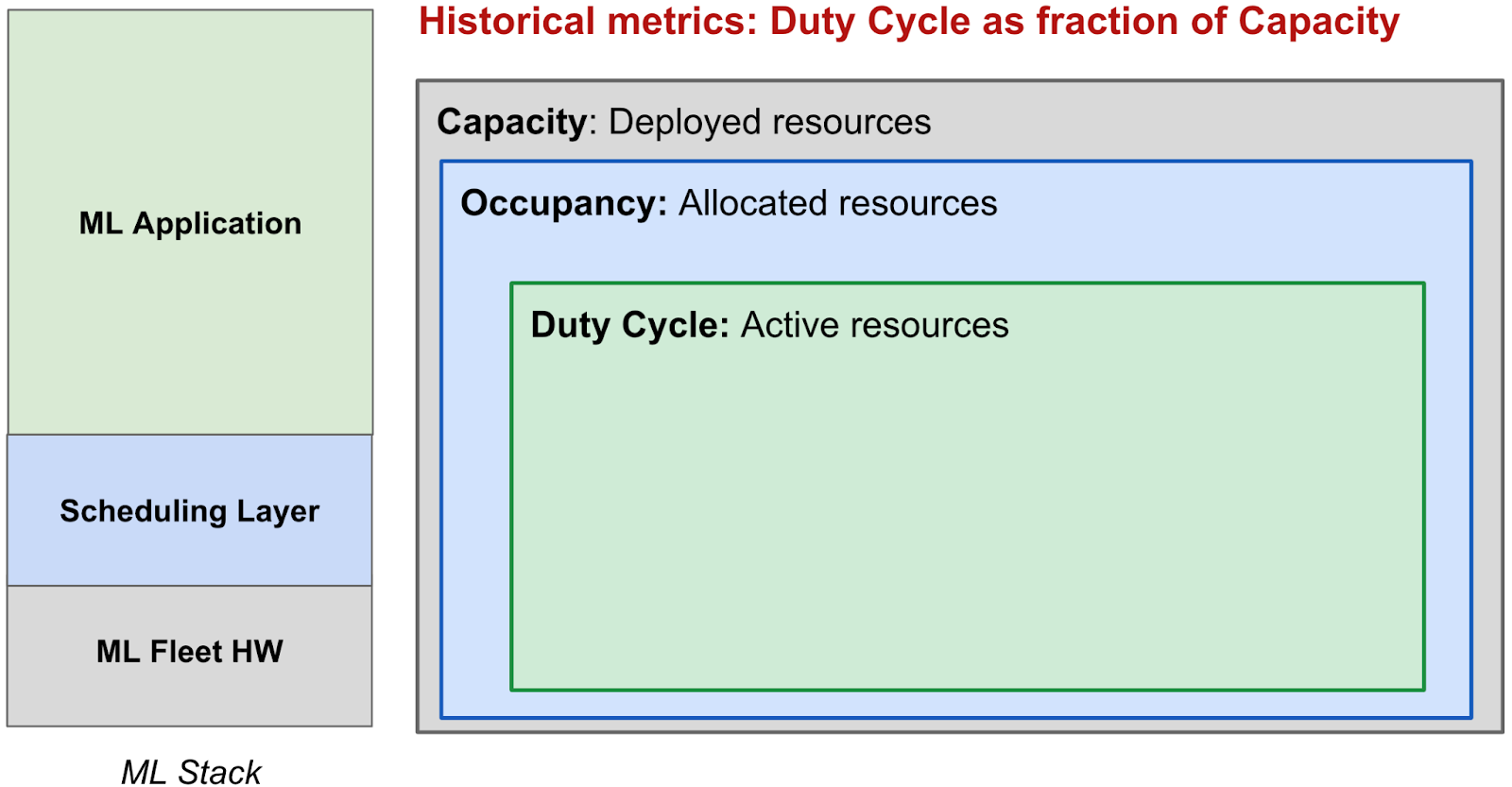}
\caption{Traditional utilization-based metrics. We replace these using goodput as a measure of fleet efficiency rather than utilization.}
\label{fig:utilization}
\end{figure}

To put things into perspective for MPG, we considered more traditional approaches to measuring ML fleet efficiency such as Capacity, Occupancy, and Duty Cycle. \autoref{fig:utilization} illustrates how we may have traditionally tended to think about performance metrics \cite{li2023analyzing, mars2011bubble,kanev2015profiling}. However, our experience reveals that these traditional metrics can sometimes fall short in providing a holistic view, given the unique challenges we have discussed in \autoref{sec:background} and \autoref{sec:fleet}. While we cannot provide specific measured details for proprietary reasons, we can share the key insights and lessons learned from our evaluation of these metrics in a production environment.

\textbf{High \underline{Capacity} equates to high resource availability.}
While capacity can tell us how many individual accelerators may be available in the fleet at a given time, it does not take into account the topological shape of those accelerators. For example, an ML training workload requesting thousands of chips in a certain physical mesh shape may never be scheduled if the only available accelerators are fragmented across different clusters or data centers. Other factors, such as the geographical location of data storage cells and accelerators, are not included in the capacity metric, even though they significantly affect job scheduling. Therefore, high capacity by itself as a metric does not necessarily effectively translate to high availability for workloads, and we should instead opt for scheduling efficiency as a more robust metric.

\textbf{High \underline{Occupancy} guarantees productivity.}
Occupancy is defined as the fraction of accelerators allocated to jobs and is often measured by the scheduler (e.g. Borg \cite{verma2015borg}). Occupancy is traditionally seen as a key efficiency indicator, but it can be misleading as it masks inefficiencies in the system stack. For example, an accelerator might be successfully allocated but stuck in I/O wait or running poorly optimized code, thus resulting in a high occupancy but very little actual progress being made towards the workload task. This is important for long-running tasks such as ML model training, where frequent pre-emptions may hinder checkpoint progress but still result in a nominally high occupancy. The traditional occupancy metric therefore does not distinguish between productive and unproductive use of allocated resources.

\textbf{\underline{Duty Cycle} accurately represents useful work.}
Duty Cycle measures whether an accelerator is in use, not how much of its compute capacity is used. When looking at an ML workload running on a TPU, duty cycle does not provide any signal on how much the matrix-multiply units (MXUs) are utilized \cite{jouppi2023tpuv4opticallyreconfigurable}. It is agnostic of the program-level efficiency and does not take into account the effectiveness of the operations being performed. An accelerator could have a high duty cycle while executing unnecessary or redundant computations. So, we require a more sophisticated metric.

\textbf{Overarching Misconception: Utilization == Productivity.}
The thread among these metrics is the assumption that keeping accelerators busy equates to productive work. However, this overlooks critical factors. (1) \textit{Quality of Computations:} None of these metrics assess whether the operations being performed are actually contributing to the desired output. (2) \textit{Workload Efficiency:} They do not consider whether the workloads are optimally designed for the hardware. (3) \textit{System-level Bottlenecks:} Focusing solely on accelerator usage ignores potential bottlenecks in data loading, memory access, or inter-accelerator communication. (4) \textit{Forward Progress:} The traditional metrics provide no insight into how much useful work is being accomplished towards completing an actual ML task, thus motivating the need for a goodput-based metric, MPG.

\section{Conclusion}

Our study presents an analysis of ML fleet efficiency using Google's TPU-based fleet. In \autoref{sec:background} and \autoref{sec:fleet}, we provide an anatomy of the ML system stack and highlight the unique challenges of performance optimization for ML fleets. These challenges include rapid model and hardware evolution and orchestration of hardware/software co-design. To address these issues, we introduce the \mpg (MPG) metric in \autoref{sec:goodput}. The MPG metric decomposes ML fleet efficiency into three key components: Scheduling, Runtime, and Program Goodputs. 

The composable nature of the MPG metric allows us to dissect efficiency trends over time and precisely identify performance bottlenecks at specific layers of the stack. Our work demonstrates that this modular approach helps us develop and deploy more targeted optimization efforts and quantify their impact in an interpretable way. In \autoref{sec:improvements}, we provide examples of real efficiency improvements from leveraging MPG with Google's own TPU fleet. The results demonstrate that the methodology presented in this paper can be generally applied to large-scale ML fleets across industry.

\section{Acknowledgements}
Our work is a product of the collaboration of many teams at Google, including the XLA team, the ML Metrics team, and the TPU Performance team. We are grateful to Daniel Herrington, Victor Cai, Kongtao Chen, Yiru Sun, Yinquan Hao, Aleksey Orekhov, Peter Ma, and Jie Sun for their support with metrics instrumentation and Amit Sabne and Shibo Wang for support with XLA optimizations. We also thank Sushma Prasad, Martin Maas, Dan Zhang, David Patterson, Michael Burrows, Charles Chang, Peter Mattson, and Michael Isard for their valuable review and feedback on this publication.

\bibliographystyle{ACM-Reference-Format}
\bibliography{refs}

\end{document}